\documentclass[sigconf]{acmart}
\AtBeginDocument{%
  }

\setcopyright{acmlicensed}
\copyrightyear{2025}
\acmYear{2025}
\acmDOI{XXXXXXX.XXXXXXX}
\acmConference[Conference acronym 'XX]{Make sure to enter the correct
  conference title from your rights confirmation email}{June 03--05,
  2018}{Woodstock, NY}
\acmISBN{978-1-4503-XXXX-X/2018/06}

\usepackage[utf8]{inputenc} 
\usepackage[T1]{fontenc}    
\usepackage{hyperref}       
\usepackage{url}            
\usepackage{booktabs}       
\usepackage{amsfonts}       
\usepackage{nicefrac}       
\usepackage{microtype}      
\usepackage{xcolor}         
\usepackage{subcaption}     
\usepackage{amsmath}        
\usepackage{multirow}       

\copyrightyear{2025}
\acmYear{2025}
\setcopyright{acmlicensed}\acmConference[ICAIF '25]{6th ACM International Conference on AI in Finance}{November 15--18, 2025}{Singapore, Singapore}
\acmBooktitle{6th ACM International Conference on AI in Finance (ICAIF '25), November 15--18, 2025, Singapore, Singapore}
\acmDOI{10.1145/3768292.3770391}
\acmISBN{979-8-4007-2220-2/2025/11}

\newcommand{\RIGHTCOMMENT}[1]{\bgroup\hfill~#1\egroup}

\begin{document}

\title{Robust time series generation via Schr\"{o}dinger Bridge: a comprehensive evaluation}

\author{Alexandre ALOUADI}
\email{alexandre.alouadi@polytechnique.edu}
\affiliation{%
  \institution{École Polytechnique \& BNP PARIBAS CIB, Global Markets}
  \city{Paris}
  \country{FRANCE}
}

\author{Baptiste BARREAU}
\email{baptiste.barreau@bnpparibas.com}
\affiliation{%
  \institution{BNP PARIBAS CIB, Global Markets}
  \city{Paris}
  \country{FRANCE}
}

\author{Laurent CARLIER}
\email{laurent.carlier@bnpparibas.com}
\affiliation{%
  \institution{BNP PARIBAS CIB, Global Markets}
  \city{Paris}
  \country{FRANCE}
}

\author{Huyên PHAM}
\email{huyen.pham@polytechnique.edu}
\affiliation{%
  \institution{École Polytechnique}
  \city{Paris}
  \country{FRANCE}
}

\renewcommand{\shortauthors}{Alouadi et al.}

\begin{abstract}
We investigate the generative capabilities of the Schr\"{o}dinger Bridge (SB) approach for time series. 
The SB framework formulates time series synthesis as an entropic optimal interpolation transport problem between a reference probability measure on path space and a target joint distribution. This results in a stochastic differential equation over a finite horizon that accurately captures the temporal dynamics of the target time series. 
While the SB approach has been largely explored in fields like image generation, there is a scarcity of studies for  its application to time series. In this work, we bridge this gap by conducting a comprehensive evaluation of the SB method's robustness and generative performance. We benchmark it against state-of-the-art (SOTA) time series generation methods across diverse datasets, assessing its strengths, limitations, and ability to model complex temporal dependencies. Our results offer valuable insights into the SB framework's potential as a versatile and robust tool for time series generation. The code is available at \url{https://github.com/alexouadi/SBTS}.
\end{abstract}




\maketitle

\section{Introduction}

Generative modeling has emerged as a powerful tool for data synthesis, with a wide range of applications in various domains, including static image processing, natural language generation, and time series modeling. The core objective  of generative modeling is to learn a probabilistic representation of the underlying data distribution, enabling the generation of  synthetic  samples that are indistinguishable from real data. Recent advances in this field have produced a variety of competing methods, each offering distinct strengths and limitations.

One prominent class of generative models is likelihood-based models, which learn the target distribution by optimizing the negative log-likelihood or its surrogate loss. 
Variational Auto-Encoders (VAEs) \cite{kingma2013autoencodingvariationalbayes} and flow-based methods \cite{dinh2015nicenonlinearindependentcomponents} are notable examples offering explicit density estimation and tractable likelihoods. While these models have demonstrated success in learning flexible distributions, they often struggle to represent complex data structures due to architectural constraints
\cite{makhzani2016adversarialautoencoders, razavi2019generatingdiversehighfidelityimages, tolstikhin2018wassersteinautoencoders, papamakarios2018maskedautoregressiveflowdensity, kingma2018glowgenerativeflowinvertible, behrmann2019invertibleresidualnetworks}.

In contrast, implicit generative models, such as Generative Adversarial Networks (GANs) \cite{goodfellow2014generativeadversarialnetworks}, 
learn to generate data by optimizing a min-max game between a generator and a discriminator. 
GANs have produced state-of-the-art results in applications like image-to-image translation and audio synthesis. However, they are notorious for training instabilities, leading to challenges like mode collapse and vanishing gradients. 
Considerable efforts have been made to improve GAN stability and performance \cite{karras2018progressivegrowinggansimproved,brock2019largescalegantraining}.

For time series generation, specialized GAN architectures have been proposed, leveraging convolutional neural networks (CNNs) \cite{Wiese_2020} and optimal transport frameworks \cite{NEURIPS2020_641d77dd}, leading to promising but still limited results \cite{NEURIPS2019_c9efe5f2}. 

Diffusion models have recently gained significant traction as a compelling alternative to GANs. These models, including Score-Based Generative Models (SGMs) \cite{song2019generativemodelingestimatinggradients} and Denoising Diffusion Probabilistic Models (DDPMs) \cite{ho2020denoisingdiffusionprobabilisticmodels}, approach generative modeling through iterative denoising. 
They add noise to data through a diffusion process and then learn to reverse this process, often using stochastic differential equations (SDEs) \cite{song2021scorebasedgenerativemodeling, tang2024scorebaseddiffusionmodelsstochastic}. SGMs utilize Langevin dynamics and neural networks, while DDPMs use Markov chains to iteratively refine noisy inputs. Diffusion models have achieved remarkable results in image synthesis and have been extended to time series generation, capturing complex temporal dynamics \cite{rasul2021autoregressivedenoisingdiffusionmodels, lim2024tsgm, lim2023regulartimeseriesgenerationusing, naiman2024utilizingimagetransformsdiffusion}.

In the specific context of time series generation, maintaining the temporal structure and statistical properties of the original data is crucial. Effective generative models must not only produce realistic individual time series but also preserve population-level characteristics, such as marginal distributions and functional dependencies across time steps. To this end, novel frameworks like the Schrödinger Bridge (SB) approach \cite{wang2021deepgenerativelearningschrodinger,Debortolietal21}, for image generation and \cite{hamdouche2023generativemodelingtimeseries} for time series  have been introduced. The SB framework formulates generative modeling as an entropic optimal transport problem, bridging a reference probability measure on path space with a target joint distribution. This results in a stochastic differential equation that inherently captures the temporal dynamics of time series data.

Despite its theoretical appeal, the SB approach remains underexplored in the time series domain, with limited empirical validation and no standardized benchmarks in the literature.

{\bf Our Contributions.} To address this gap, we conduct a comprehensive evaluation of the Schr\"odinger Bridge method for time series generation. Specifically, we:
\begin{itemize}
\item Benchmark the SB approach against state-of-the-art (SOTA) generative models for time series using their established metrics.
\item Introduce new evaluation metrics to better assess the quality and robustness of generated time series, focusing on temporal dependencies and statistical fidelity.
\item Propose an improvement to the existing method for long time series generation, along with a practical solution to the critical issue of hyperparameter selection.

\end{itemize}
This study provides the first systematic evaluation of the Schr\"odinger Bridge approach for time series generation, offering valuable insights into its practical utility and potential for future applications.

\section{Background}

\subsection{Classical Schr\"{o}dinger Bridge problem}

We recap the formulation of  the classical Schr\"{o}dinger Bridge problem (\textbf{SBP}) for two marginals constraints, see \cite{leo14, chenetal21}.  

We denote by  $\Omega = C([0,T],\mathbb{R}^d)$ the space of continuous $\mathbb{R}^d$-valued paths on $[0, T], T < \infty$, $X = (X_t)_{t\in [0,T]}$ the canonical process, i.e. $X_t(\omega) = \omega_t$, $\omega = (\omega_s)_{s\in [0,T]}\in \Omega$. Let $\mathcal{P}(\Omega)$ be the set of probability measures on path space $\Omega$. For $\mathbb{P} \in \mathcal{P}(\Omega)$, $\mathbb{P}_t = X_t \# \mathbb{P} = \mathbb{P} \circ X^{-1}_t$, is the marginal law of $X_t$. In other words,  $\mathbb{P}_t$ is the law of the particle at time $t$, when the law of the whole trajectory is $\mathbb{P}$. 

Let $\mu_0$ and $\mu_T$ be two probability measures on $\mathbb{R}^d$, and $\mathbb{Q}$ be a prior/reference measure on $\Omega$, which represents the belief of the dynamics before data observation, e.g., the law of Wiener process with initial measure $\nu_0$.

The  \textbf{SBP} can be formulated as follows: Find a measure $\mathbb{P}^*$ on path space $\mathcal{P}(\Omega)$  such that
\[
\mathbb{P}^*  \in {\arg\min}_\mathbb{P} \{ \text{KL}(\mathbb{P}|\mathbb{Q}): \mathbb{P} \in \mathcal{P}(\Omega), \mathbb{P}_0 = \mu_0, \mathbb{P}_T=\mu_T \},
\]
where $\text{KL}(\mathbb{P}|\mathbb{Q}) := \int \log\left(\frac{d\mathbb{P}}{d\mathbb{Q}}\right) d\mathbb{P}$ if $\mathbb{P} \ll \mathbb{Q}$, else $\infty$, is the Kullback-Leibler divergence (or relative entropy) between two nonnegative measures. 

Introducing $\mathbb{P}_{0,T} = \mathbb{P} \circ (X_0, X_T)^{-1}$ the joint initial-terminal law of $(X_0, X_T)$ under $\mathbb{P}$, we then have $\mathbb{P}[\cdot] = \nobreak \int \mathbb{P}[\cdot] ^{xy} \mathbb{P}_{0,T}(dx, dy)$ with $\mathbb{P}[\cdot]^{xy} = \mathbb{P}[\cdot | (X_0, X_T) = \nobreak (x,y)]$, and similarly for $\mathbb{Q}, \mathbb{Q}^{xy}$ and $\mathbb{Q}_{0,T}$. Now using that ${\text{KL}}(\mathbb{P} | \mathbb{Q}) = {\text{KL}}(\mathbb{P}_{0,T} | \mathbb{Q}_{0,T}) + \iint {\text{KL}}(\mathbb{P}^{xy} | \mathbb{Q}^{xy}) \mathbb{P}_{0,T}(dx, dy)$, one can reduce the \textbf{SBP} to a static SB by minimizing :
\[
\text{KL}(\pi|\mathbb{Q}_{0,T}) = \iint \log(\frac{d\pi}{d\mathbb{Q}_{0,T}}(x,y))(dx, dy)
\]
over the couplings $\pi \in  \Pi(\mu_0, \mu_T) = \{ \pi \in \mathcal{P}(\mathbb{R}^d \times \mathbb{R}^d), \pi_0 = \mu_0, \pi_T=\mu_T\}$.

The solution of the dynamic \textbf{SBP} is then given by $\mathbb{P}^* = \int \mathbb{Q}^{xy} \pi^* (dx,dy)$, where $\pi^*$ is solution to the static \textbf{SBP}. 

Let us now consider the case where $\mathbb{Q} = \mathbb{W}^{\sigma}$ the Wiener measure of variance $\sigma^2$, i.e., the law of the process $X_t = X_0 + \sigma W_t, 0 \leq t \leq T, X_0 \sim \nu_0$, with $W$ a Brownian motion. 
If $\mathbb{P} \in \mathcal{P}(\Omega)$ such that KL$(\mathbb{P}|\mathbb{W}^{\sigma}) < \infty$, there exists an $\mathbb{R}^d$-valued process $\alpha$, adapted w.r.t $\mathbb{F}$ the canonical filtration, with $\mathbb{E}_{\mathbb{P}} [ \int_0^T |\frac{\alpha_t}{\sigma}|^2 dt] < \infty$ such that 
\[
\frac{d\mathbb{P}}{d\mathbb{W}^{\sigma}} = \frac{d\mathbb{P}_0}{d\nu_0} \exp(\int_0^T \frac{\alpha_t}{\sigma}dW_t^{\mathbb{P}} + \frac{1}{2} \int_0^T \left \lVert {\frac{\alpha_t}{\sigma}} \right \rVert ^2 dt)
\]
and by Girsanov's theorem, under $\mathbb{P}$, $dX_t = \nobreak \alpha_t dt + \sigma dW_t^{\mathbb{P}}, 0 \leq t \leq T$ with $W^{\mathbb{P}}$ a Brownian motion under $\mathbb{P}$. At the end, we have KL$(\mathbb{P}|\mathbb{W}^{\sigma}) = \text{KL}(\mathbb{P}_0|\nu_0) + \mathbb{E}_{\mathbb{P}} [ \int_0^T |\frac{\alpha_t}{\sigma}|^2 dt]$. We then can reformulate the \textbf{SBP} problem as a stochastic control problem over the drift $\alpha$ as follows:

\begin{equation}\label{e:eq_sb} 
	\min_{\alpha} \mathbb{E}_{\mathbb{P}} \left[ \int_0^T\left \lVert {\frac{\alpha_t}{\sigma}} \right \rVert ^2 dt\right]
\end{equation}

such that  $dX_t = \nobreak \alpha_t dt + \sigma dW_t^{\mathbb{P}}, X_0 \sim \mu_0, X_T \sim \mu_T$.

\subsection{Schr\"{o}dinger Bridge problem for time series}

We now formulate the Schr\"{o}dinger Bridge problem for time series introduced in \cite{hamdouche2023generativemodelingtimeseries}. 
Let $\mu$ be the distribution of a time series valued in $\mathbb{R}^d$ of which we can observe samples over a discrete time grid $\mathrm{T} = \{t_1, \cdots, t_N = T\}$. We want to construct a model capable of generating time series samples that follow the distribution $\mu \in \mathcal{P}((\mathbb{R}^{d})^N)$ given real observations.

The \textbf{SBP} \eqref{e:eq_sb} for time series generation, noted Schr\"{o}dinger Bridge Time Series (\textbf{SBTS}) is formulated as follows: 
\begin{equation}\label{e:eq_sbts}
	\min_{\alpha} \mathbb{E}_{\mathbb{P}} \left[ \int_0^T \left \lVert \alpha_t \right \rVert^2 dt\right]
\end{equation}

such that  $dX_t = \alpha_t dt +   dW_t^{\mathbb{P}}$ with $W$ a Brownian motion under $\mathbb{P}$, $X_0 = \textbf{0}, (X_{t_1}, \cdots, X_{t_N}) \overset{\mathbb{P}}{\sim} \mu$. 
Note that we considered the case $\sigma = 1$ to simplify the notation, without loss of generality.

\begin{theorem}\label{th01} 
\cite{hamdouche2023generativemodelingtimeseries}
The diffusion process 
\[
X_t = \int_0^t \alpha^*_s \, ds + W_t, \quad 0 \leq t \leq T,
\]
with $\alpha^*$ defined as
\[
\alpha^*_t = a^*(t, X_t; \textbf{X}_{\eta(t)}), \quad 0 \leq t < T,
\]
solves the \textbf{SBTS} problem \eqref{e:eq_sbts}, with 
\[
\eta(t) = \max\{t_i : t_i \leq t\},
\]
and
\begin{equation}
a^*(t, x; \textbf{x}_i) =
\frac{1}{t_{i+1} - t}
\frac{\mathbb{E}_\mu \left[ (X_{t_{i+1}} - x)\,
F_i(t, X_{t_i}, x, X_{t_{i+1}}) \,\middle|\, \textbf{X}_{t_i} = \textbf{x}_i \right]}
{\mathbb{E}_\mu \left[ F_i(t, X_{t_i}, x, X_{t_{i+1}}) \,\middle|\, \textbf{X}_{t_i} = \textbf{x}_i \right]} 
\label{eq:a_star_equation}
\end{equation}

for $t \in [t_i, t_{i+1})$, 
$\textbf{x}_i = (x_1, \ldots, x_i) \in (\mathbb{R}^d)^i$, $x \in \mathbb{R}^d$, where
\[
F_i(t, x_i, x, x_{i+1}) = \exp \left\{
- \frac{\lVert x_{i+1} - x \rVert^2}{2(t_{i+1} - t)} 
+ \frac{\lVert x_{i+1} - x_i \rVert^2}{2(t_{i+1} - t_i)} 
\right\}.
\]
\end{theorem}

To estimate the drift, one can employ a kernel density estimation method using $M$ data samples $\boldsymbol{X}_{t_N}^m = (X^m_{t_1}, \cdots, X^m_{t_N}), m = 1, \cdots, M$:

\begin{equation}\label{e:a_hat}
	\hat{a}(t, x; \boldsymbol{x}_i) =\frac{1}{t_{i+1} - t} \frac{\sum_{m=1}^{M} (X^{(m)}_{t_{i+1}} - x) F_i(t, X^{(m)}_{t_i}, x, X^{(m)}_{t_{i+1}})\tilde{K}_i^m}{\sum_{m=1}^{M} F_i(t, X^{(m)}_{t_i}, x, X^{(m)}_{t_{i+1}}) \tilde{K}_i^m}
\end{equation}
with $\tilde{K}_i^m = \prod_{j=1}^{i} K_h(x_j - X^{(m)}_{t_j})$ for $t \in [t_i, t_{i+1}[$, $\boldsymbol{x}_i \in (\mathbb{R}^d)^i$, $i \in \{1, \cdots N-1\}$ and $K_h$ a kernel function, defined here as $K_h(x) = \frac{1}{h^d}(1 - \lVert {\frac{x}{h}} \rVert^2)^2 1_{\lVert {x} \rVert < h}$ with bandwidth $h >0$. We emphasize that this approach  does not necessitate any pre-training. The methodology is straightforward, as it involves the incremental construction of a sample, leveraging a deterministic drift estimate.

\section{Bandwidth selection and long series generation}
\label{sec:select_k_h}

The choice of the bandwidth $h$ is critical for our generated data. On one hand, if $h$ is too small, the estimation will be too rough, capturing noise and leading to high variance. On the other hand, if $h$ is too large, the estimation introduces a high bias and may fail to adapt to changes in the data. There have been many studies in the literature on bandwidth selection. The most popular ones are rule-of-thumb and cross-validation methods \cite{article}, but we found these approaches not useful for our problem.

\vspace{-0.5em}

\begin{figure}[ht]
	\begin{center}
		\centerline{\includegraphics[width=1\columnwidth]{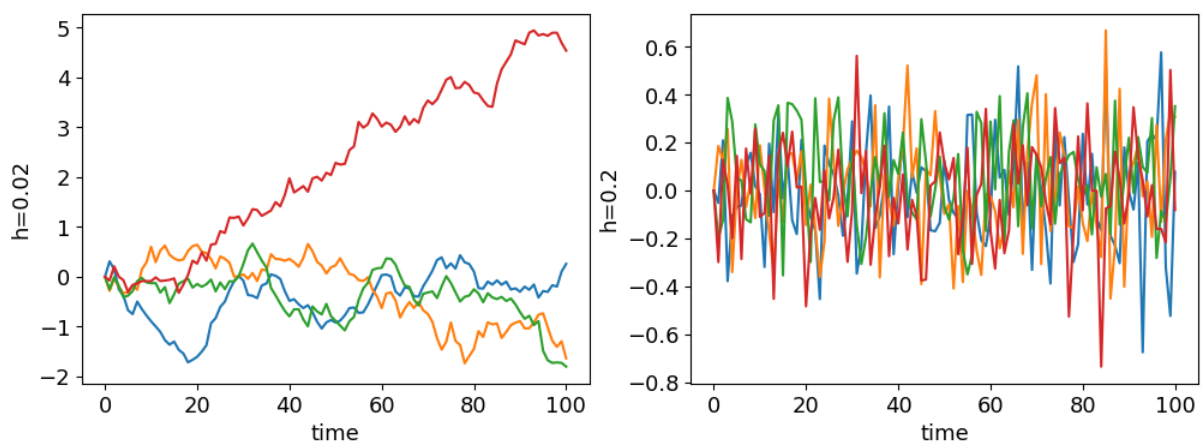}}
		\caption{Generation of a Markovian GARCH model of order 2 (see \autoref{sec:appendix_dataset}), with $h=0.02$ (\textit{left}) and $h=0.2$ (\textit{right}). We can clearly see the effect of a too small $h$, here $0.02$, as the left plot exhibits undesirable behavior, whereas the right one displays the desired outcome.}
		\label{fig:h_selection}
	\end{center}
	\vskip -0.1in
\end{figure}

We then propose a simple approach to select the bandwidth, by considering it as a hyper-parameter to fine-tune. Given a train set  $X = (X^m_{t_1}, \cdots, X^m_{t_N})_{m=1, \cdots, M}$, a test set $Y = (Y^q_{t_1}, \cdots, Y^q_{t_N})_{q=1, \cdots, Q}$, both from real data, and a list of bandwidths $H = \{h_1, \cdots, h_K\}$, we generate $L$ realizations of $\hat{Y}^q_{t_N}$ given the first real values of the series $(Y^q_{t_1}, \cdots, Y^q_{t_{N-1}})$ for each $q$, using \eqref{e:a_hat}.
Then, we choose $h^* \in H$ such that it minimizes
\begin{equation}\label{e:mse_h}
	MSE_h = \frac{1}{Q} \sum_{q=1}^Q \left\lvert \frac{1}{L} \sum_{l=1}^L \hat{Y}_{t_N}^{q,l} - Y_{t_N}^q \right\rvert^2
\end{equation}
with $\hat{Y}_{t_N}^{q,l}$ the $l$-th generated realizations of $\hat{Y}^q_{t_N}$ given $(Y^q_{t_1}, \cdots, Y^q_{t_{N-1}})$.

Moreover, for long time series, it is most likely that $\tilde{K}_i^m$ becomes null when $i$ is large enough for each $m$, because of the condition $1_{\lVert {x_j - X_{t_j}} \rVert < h}$. In that case, the drift cannot be properly estimated, and as a result the generated sample may be inaccurate. One solution would be to increase $h$, which doesn't seem optimal since $h$ should be as small as possible, due to the bias-variance trade-off. To address this issue, one may assume the series to be Markovian of order $k$, and use only $(X_{t_j}^m, x_{t_j})_{j= i-k+1, \cdots, i}, m = 1, \cdots, M$ to generate $x_{t_{i+1}}$, replacing $\tilde{K}_i^m = \prod_{j=1}^{i} K_h(x_j - X^{(m)}_{t_j})$ by $\tilde{K}_{k,i}^m = \prod_{j=i-k+1}^{i} K_h(x_j - X^{(m)}_{t_j})$. We can select the parameter $k$ the same way as $h$ in parallel.

\begin{figure}[ht]
	\begin{center}
		\centerline{\includegraphics[width=.8\columnwidth]{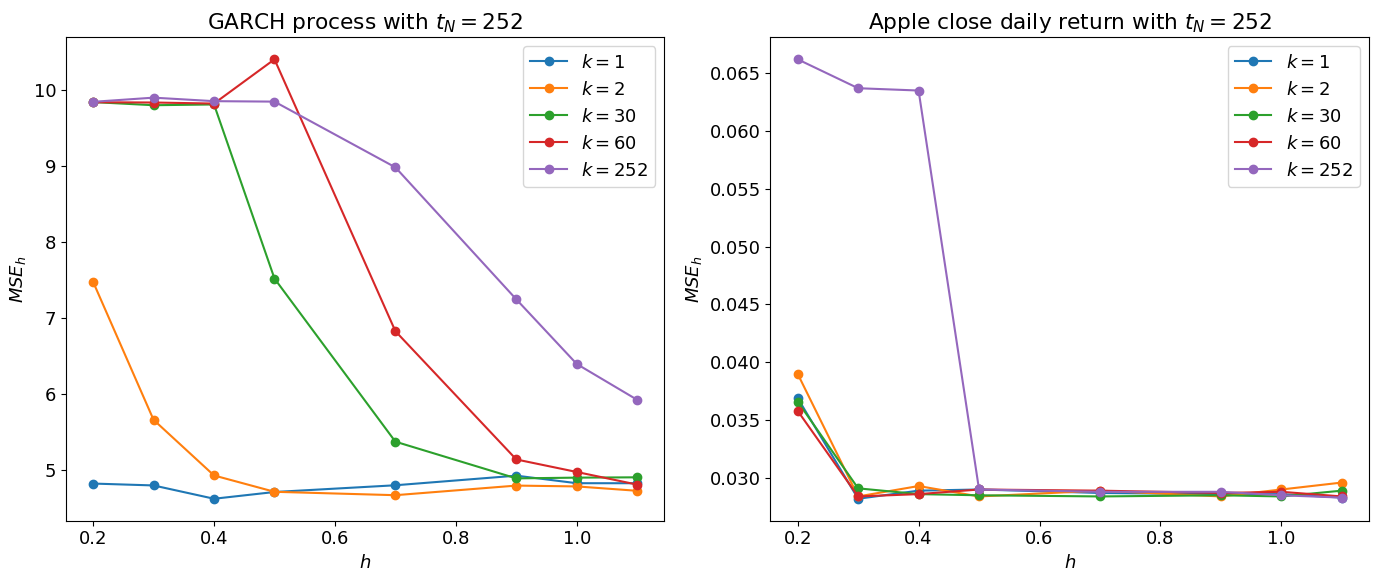}}
			\caption{Bandwidth selection for GARCH data defined as in \autoref{sec:appendix_dataset} and Apple close daily return of length $252$, using different values of $k$. For both time series, we see that a larger $h$ is needed when using the whole time series ($k=252$). Otherwise, a smaller $h$ is sufficient, and the Markovianity even improved the $MSE$ for GARCH data, as it is really Markovian. } 
		\label{fig:demo}
	\end{center}
	\vskip -0.3in
\end{figure}

Note that for multivariate time series, it is possible to employ feature-specific bandwidths, as the scale and magnitude of each feature may vary. In that case, we should replace the previous kernel $K_h(x) = \frac{1}{h^d}(1 - \| \frac{x}{h} \|^2)^2 1_{\lVert {x} \rVert < h}$ by $K_h(x) = \prod_{k=1}^d \frac{1}{h_k}(1 - |\frac{x_k}{h_k}|^2)^2 1_{|x_k| < h_k}$ with $d$ the number of time series.

\section{SBTS vs SOTA}

In this section, we compare the model with SOTA approaches across multiple datasets. We analyze the performance of the method relative to the results obtained on these datasets to evaluate its effectiveness and robustness. 

\subsection{Datasets}
\label{sec:dataset_bench}

We performed our tests using three real-world and two toy datasets, all multivariates:

\begin{itemize}
\setlength\itemsep{-0.2em}
\item \textbf{Stock} \cite{NEURIPS2019_c9efe5f2}, consisting in Google daily historical data from 2004 to 2019, including 6 features: high, low, opening, closing, adjusted closing prices and volume. 
\item \textbf{Energy} \cite{article_candanedo}, noisy energy consumption data from the UCI repository including 28 features, e.g., temperature or wind speed.
\item \textbf{Air} \cite{article_de_vito}, from the UCI repository, containing gas sensor readings and reference concentrations from an Italian city, averaged hourly including 13 features.
\item \textbf{Multi-Stocks} \cite{tao2024highrankpathdevelopment}, includes daily log return of 5 representative
stocks in the U.S. market from 2010 to 2020
\item \textbf{Sine} \cite{NEURIPS2019_c9efe5f2}, sine wave data of dimension 5
\item \textbf{AR} \cite{NEURIPS2019_c9efe5f2}, auto-regressive Gaussian model of dimension 5, with parameters $\phi = 0.5$ and $\sigma = 0.8$.
\item \textbf{fBM} \cite{tao2024highrankpathdevelopment}, fractional Brownian Motion with $H=\frac{1}{4}$

\end{itemize}

For this experiment, we used time series of length $24$, in line with SOTA methods. For more details on the definition of the Sine and AR datasets, refer to \autoref{sec:appendix_dataset}.

\subsection{Evaluation Metrics}

We will refer in this section to two widely used metrics. \textbf{1) Discriminative score: } \cite{NEURIPS2019_c9efe5f2} a classifier is trained to output the probability of a given sequence being real. Then, we compute the accuracy on an equally composed test set of both real and synthetic data. We aim to minimize the following discriminative score: $score = |acc - 0.5|$ where $acc$ is the accuracy on the test set. Note that we target an accuracy close to $0.5$, as it indicates that the classifier is unable to distinguish between real and synthetic data, effectively resorting to random guessing.  \textbf{2) Predictive score: } \cite{NEURIPS2019_c9efe5f2} a model is trained on synthetic data only, to predict the next data point in a given time series, and test on real data only. We aim to minimize the global mean absolute error, which measures the average difference between predicted and actual values.

We also used the Auto and Cross-Correlation on the Multi-Stock and fBM datasets, along with the Outgoing Nearest Neighbour Distance score (ONDD) \cite{leznik2022sok}.

\subsection{Results}
\label{sec:benchmark_res}

\begin{table}[h]
\caption{Comparative results highlighting discriminative and predictive scores. The highest scores are denoted in bold. N/A indicates that the value is not available.}
\label{tab:real_data}
\centering
\small
\begin{tabular}{lccc}
\hline
\textbf{Method} & \textbf{Stocks} & \textbf{Energy} & \textbf{Air} \\
\hline
\multicolumn{4}{l}{\textit{Disc. Score}} \\
TSGM-VP & .022$\pm$.005 & .221$\pm$.025 & .122$\pm$.014 \\
TSGM-subVP & .021$\pm$.008 & .198$\pm$.025 & .127$\pm$.010 \\
ImagenTime & .037$\pm$.006 & \textbf{.040}$\pm$\textbf{.004} & N/A \\
T-Forcing & .226$\pm$.035 & .483$\pm$.004 & .404$\pm$.020 \\
P-Forcing & .257$\pm$.026 & .412$\pm$.006 & .484$\pm$.007 \\
TimeGAN & .102$\pm$.031 & .236$\pm$.012 & .447$\pm$.017 \\
RCGAN & .196$\pm$.027 & .336$\pm$.017 & .459$\pm$.104 \\
C-RNN-GAN & .399$\pm$.028 & .499$\pm$.001 & .499$\pm$.000 \\
TimeVAE & .175$\pm$.031 & .498$\pm$.006 & .381$\pm$.037 \\
WaveGAN & .217$\pm$.022 & .363$\pm$.012 & .491$\pm$.013 \\
COT-GAN & .285$\pm$.030 & .498$\pm$.000 & .423$\pm$.001 \\
SBTS & \textbf{.010 $\pm$ .008} & .356 $\pm$ .020 & \textbf{.036 $\pm$ .016} \\
\hline
\multicolumn{4}{l}{\textit{Pred. Score}} \\
TSGM-VP & .037$\pm$.000 & .257$\pm$.000 & \textbf{.005$\pm$.000} \\
TSGM-subVP & .037$\pm$.000 & .252$\pm$.000 & \textbf{.005$\pm$.000} \\
ImagenTime & .036$\pm$.000 & .250$\pm$.000 & N/A \\
T-Forcing & .038$\pm$.001 & .315$\pm$.005 & .008$\pm$.000 \\
P-Forcing & .043$\pm$.001 & .303$\pm$.006 & .021$\pm$.000 \\
TimeGAN & .038$\pm$.001 & .273$\pm$.004 & .017$\pm$.004 \\
RCGAN & .040$\pm$.001 & .292$\pm$.005 & .043$\pm$.000 \\
C-RNN-GAN & .038$\pm$.000 & .483$\pm$.005 & .111$\pm$.000 \\
TimeVAE & .042$\pm$.002 & .268$\pm$.004 & .013$\pm$.002 \\
WaveGAN & .041$\pm$.001 & .307$\pm$.007 & .009$\pm$.000 \\
COT-GAN & .044$\pm$.000 & .260$\pm$.000 & .024$\pm$.001 \\
SBTS & \textbf{.017 $\pm$ .000} & \textbf{.072 $\pm$ .001} & \textbf{.005 $\pm$ .001} \\
\hline
\end{tabular}
\end{table}

\begin{table}[h]
\caption{Results on toy datasets: Sine, AR.}
\label{tab:toy_data}
\centering
\small
\begin{tabular}{lcc}
\hline
\textbf{Method} & \textbf{Sine} & \textbf{AR} \\
\hline
\multicolumn{3}{l}{\textit{Disc. Score}} \\
T-Forcing & .495$\pm$.001 & .500$\pm$.000 \\
P-Forcing & .430$\pm$.027 & .472$\pm$.008 \\
ImagenTime & .014$\pm$.009 & N/A \\
TimeGAN & \textbf{.011$\pm$.008} & .174$\pm$.012 \\
RCGAN & .022$\pm$.008 & .190$\pm$.011 \\
C-RNN-GAN & .229$\pm$.040 & .227$\pm$.017 \\
WaveNet & .158$\pm$.011 & .235$\pm$.009 \\
WaveGAN & .277$\pm$.013 & .213$\pm$.013 \\
SBTS & .061 $\pm$ .010 & \textbf{.034 $\pm$ .003} \\
\hline
\multicolumn{3}{l}{\textit{Pred. Score}} \\
T-Forcing & .150$\pm$.022 & .732$\pm$.012 \\
P-Forcing & .116$\pm$.004 & .571$\pm$.005 \\
ImagenTime & .094$\pm$.000 & N/A \\
TimeGAN & \textbf{.093$\pm$.019} & .412$\pm$.002 \\
RCGAN & .097$\pm$.001 & .435$\pm$.002 \\
C-RNN-GAN & .127$\pm$.004 & .490$\pm$.005 \\
WaveNet & .117$\pm$.008 & .508$\pm$.003 \\
WaveGAN & .134$\pm$.013 & .489$\pm$.001 \\
SBTS & .095 $\pm$ .002 & \textbf{.092 $\pm$ .007} \\
\hline
\end{tabular}
\end{table}

\begin{table}[h]
\caption{Additional results for fBM and Stock datasets.}
\label{tab:additional_results}
\centering
\small
\begin{tabular}{llcccc}
\hline
\textbf{Dataset} & \textbf{Metric} & \textbf{RCGAN} & \textbf{TimeGAN} & \textbf{PCFGAN} & \textbf{SBTS} \\
\hline
\multirow{5}{*}{fBM} 
& Auto-C. & .105$\pm$.001 & .459$\pm$.003 & .125$\pm$.003 & \textbf{.017$\pm$.004} \\
& Cross-C. & .051$\pm$.001 & .092$\pm$.001 & .047$\pm$.001 & \textbf{.005$\pm$.000} \\
& Disc. score & .207$\pm$.008 & .480$\pm$.002 & .265$\pm$.006 & \textbf{.005$\pm$.005} \\
& Pred. score & .456$\pm$.004 & .686$\pm$.013 & .474$\pm$.003 & \textbf{.423$\pm$.000} \\
& ONND & .622$\pm$.002 & .632$\pm$.002 & .654$\pm$.002 & \textbf{.471$\pm$.004} \\
\hline
\multirow{5}{*}{Stock}
& Auto-C. & .239$\pm$.016 & .228$\pm$.010 & .198$\pm$.003 & .192$\pm$.008 \\
& Cross-C. & .067$\pm$.011 & .056$\pm$.002 & .055$\pm$.004 & \textbf{.032$\pm$.001} \\
& Disc. score & .134$\pm$.058 & .020$\pm$.021 & .028$\pm$.017 & \textbf{.012$\pm$.010} \\
& Pred. score & .010$\pm$.000 & .009$\pm$.000 & .009$\pm$.000 & \textbf{.008$\pm$.000} \\
& ONND & .017$\pm$.001 & .017$\pm$.000 & \textbf{.016$\pm$.000} & .025$\pm$.000 \\
\hline
\end{tabular}
\end{table}

Table  \ref{tab:real_data}, \ref{tab:toy_data} and \ref{tab:additional_results} show that SBTS outperforms most SOTA models, including those based on GAN approaches. However, SBTS either performs similarly to TSGM and ImagenTime models \cite{lim2024tsgm, naiman2024utilizingimagetransformsdiffusion}, which are diffusion-based models for time series or outperforms it depending on the dataset. A notable exception is the Energy dataset, for which the discriminative score is not directly comparable to the best-performing model; this is due to the presence of abrupt jumps in the data, which are not ideally suited to the model based on SDE that assumes smoother dynamics. Moreover, it is worth mentioning that the SBTS approach is faster, taking at most a couple of hours to generate all the samples, and requires no hyperparameter fine-tuning, except for $h$, and $k$ if the series is long. 
  
Note that the method requires stationary time series data, which we achieved - if needed - by transforming the data into log returns for simulation, and then converting them back on base one scale (see more details in \autoref{sec:data_scaling}). However, SOTA models generate data that are normalized using min-max scaling, therefore, we also applied min-max scaling to our data to ensure a comparable scale while computing the scores.

\begin{figure}
    \centering
    \begin{subfigure}[t]{0.48\columnwidth}
        \centering
        \includegraphics[width=\linewidth]{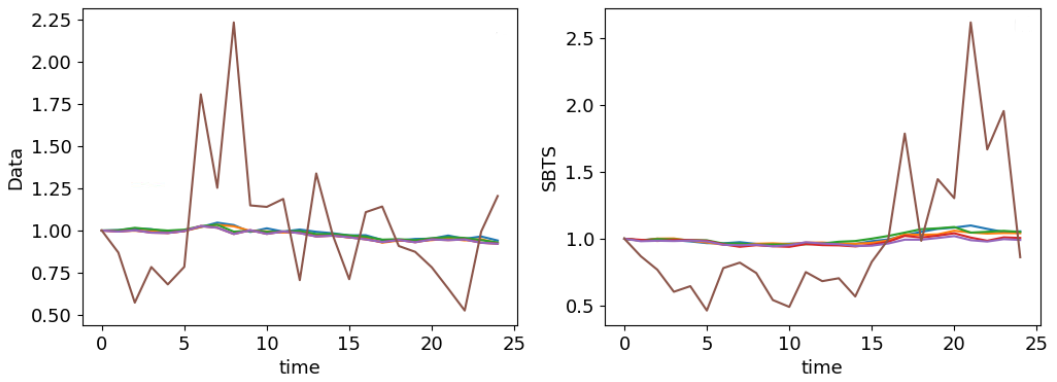}
        \caption{Comparison between original data (left) and SBTS (right) for \textbf{Stock} data on base-one scale. A random subset is selected and visualized. Volume is shown in brown, other features in distinct colors.}
        \label{fig:stock_sbts}
    \end{subfigure}
    \hfill
    \begin{subfigure}[t]{0.48\columnwidth}
        \centering
        \includegraphics[width=\linewidth]{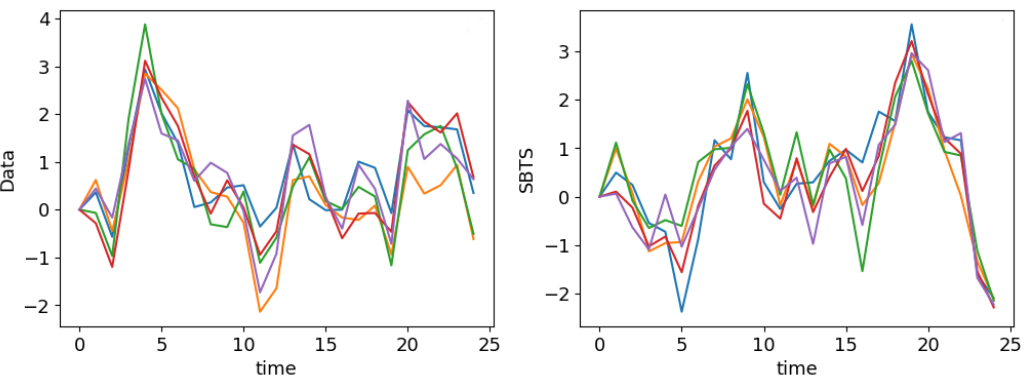}
        \caption{Comparison between original data (left) and SBTS (right) for AR data. A random subset is selected and visualized.}
        \label{fig:ar_sbts}
    \end{subfigure}
    \caption{Comparison of original and SBTS data for Stock and AR datasets.}
    \label{fig:sbts_comparison}
    \vskip -0.1in
\end{figure}

\section{Additional robustness test}

\subsection{Framework}

To evaluate the robustness of the generative model, we designed an experimental framework involving the simulation of parametric stochastic processes and the estimation of their underlying parameters from both real-world data and synthetic data generated by the model. For each real-world sample, we employed a uniform sampling strategy to randomly select parameters from a predefined range. The primary objective of this assessment is to determine whether the model can effectively capture a ``distribution of distributions''. The distributions of estimated parameters obtained from real-world and simulated data. To infer the parameters, we utilize the Maximum Likelihood Estimation (MLE) method (detailed below). Specifically, we investigated two well-established stochastic processes: 

\begin{itemize}
	\item \textbf{Ornstein-Uhlenbeck Process: } This process is defined as :
	\[
	dX_t = \theta (\mu - X_t) dt + \sigma dW_t
	\]
	where $\theta > 0, \sigma > 0, \mu$ are the parameters and $W_t$ a Brownian motion. It is well-known that $X_{t+\Delta t} | X_t \sim \nobreak \mathcal{N}(\mu_t,\sigma_t^2)$ with 

	\[
	\begin{cases}
	\mu_t = X_t e^{-\theta \Delta t} + \mu(1 - e^{-\theta \Delta t}) \\ 
	\sigma_t^2 =  \frac{\sigma^2}{2\theta}\left(1 - e^{-2\theta \Delta t}\right)
	\end{cases}
	\] 
    
    To generate a sample, one can use the following discretization:
    
    \[ X_{t + \Delta t}\! = \mu_t + \sigma_t \mathcal{N}(0,1)
    \]
    Then, for each sample, the negative log-likelihood to minimize is : 
    
    \[\mathcal{L}_m (\theta, \mu, \sigma; X^m) = - \sum_{i=1}^{N-1}  \left[-\frac{1}{2} \log(2 \pi {\sigma^2_{t_i}}^m) - \frac{(X^m_{t_{i+1}} - \mu_{t_i}^m)^2}{2  {\sigma^2_{t_i}}^m} \right]
    \] for $m =1, \cdots, M$, where $\mu_{t_i}^m$ and ${\sigma^2_{t_i}}^m$ refer to the mean and the variance of the $m$-th sample. 
	
	\item \textbf{Heston Process: } This $2$-dimensional process is defined as :
	\[
	\begin{cases}
	dX_t = r X_t dt + \sqrt{v_t} X_t dW_t^X \\ 
	dv_t = \kappa(\theta - v_t)dt + \xi \sqrt{v_t}dW_t^v 
	\end{cases}
	\] 
	where $\kappa > 0, \theta > 0, \xi > 0, r, \rho=Cor(W_t^X,W_t^v) \in [-1, 1]$ are the parameters. 
    Using Itô's lemma, one can show that \\
    $X_{t+ \Delta t}\! =\! X_t \exp\left( \int_{t}^{t+ \Delta t} (r - \frac{v_u}{2}) du +\! \int_{t}^{t+ \Delta t}\! \sqrt{v_u}dW_u^X \right)$
    
    However, for a small $\Delta t$, one can assume that 
    \[
    \int_{t}^{t+ \Delta t} (r - \frac{v_u}{2}) du \simeq (r - \frac{v_t}{2}) \Delta t,
    \]
    as well as 
    \[
    \int_{t}^{t+ \Delta t} \sqrt{v_u}dW_u^X  \simeq  \sqrt{v_t} (W_{t + \Delta t} - W_t).
    \]
    Using that $(W_{t + \Delta t} - W_t) \sim \mathcal{N}(0, \Delta t)$, we can define $Y_t$ such that
    \[
    Y_t = 
    \begin{pmatrix}
        \log\left(\frac{X_{t + \Delta t}}{X_t}\right) \\
        v_{t + \Delta t} - v_t
    \end{pmatrix}
    \sim \mathcal{N} \left(
    \mu_t,
    \Sigma_t
    \right).
    \]

    with 
    \[
    \mu_t = 
    \begin{pmatrix}
        \mu_t^X \\
        \mu_t^v
    \end{pmatrix} = 
    \begin{pmatrix}
        (r - \frac{1}{2} v_t) \Delta t \\
        \kappa (\theta - v_t) \Delta t
    \end{pmatrix}
    \]
    and 
    \[
    \Sigma_t  =
    \begin{pmatrix}
        v_t \Delta t & \rho \xi v_t \Delta t \\
        \rho \xi v_t \Delta t & \xi^2 v_t \Delta t
    \end{pmatrix}.
    \]
    
    Now, we can use the following discretization, similarly to the previous case: 
    \[
    \begin{cases}
        X_{t+ \Delta t} = X_t \exp\left( \mu_t^X + \sqrt{v_t \Delta t} Z_1 \right) \\
        v_{t+ \Delta t} = v_t + \mu_t^v +  \xi \sqrt{v_t \Delta t} Z_2
    \end{cases}
    \]

    with \((Z_1, Z_2) \sim \mathcal{N} \left( \begin{pmatrix}
        0 \\
        0
    \end{pmatrix},
    \begin{pmatrix}
        1 & \rho \\
        \rho & 1
    \end{pmatrix}\right)\), and minimize the following negative log-likelihood: \\

    $\mathcal{L}_m(\kappa, \theta, r, \rho, \xi ; \{X^m, v^m\}) = $

\begin{align*}
- \sum_{i=1}^{N-1} \left[
   -\tfrac{1}{2} \log\!\big(4\pi^2|\Sigma_{t_i}^m|\big)
   - \tfrac{1}{2} (Z_{t_i}^m)^{\top} (\Sigma_{t_i}^m)^{-1} Z_{t_i}^m
   \right]
\end{align*}

 for $m = 1, \cdots, M$, with $Z_{t} = (Y_{t} - \mu_{t})$ and $|\Sigma_{t}| = \det(\Sigma_t)$.
    
\end{itemize} 

\subsection{Results}
\label{sec:res_5_2}

In all our experiments, we generated $1000$ time series of length $252$ for both the Ornstein-Uhlenbeck and Heston processes, with $\Delta t_i = \frac{1}{252}$, after performing bandwidth and Markovian order selection, as discussed in \autoref{sec:select_k_h}. To mitigate the impact of outliers, we present the results for the parameters within the  $1^{th}$ and  $99^{th}$ percentile range, thereby providing a more robust representation of the data distribution in the plots. We denote by \textit{Data} the real data samples, \textit{SBTS} the synthetic data generated using SBTS, and \textit{Real} the range from which we randomly select the parameters for each data sample. Additional information can be found in \autoref{sec:appendix}.

\begin{figure}[h]
	\begin{center}
		\centerline{\includegraphics[width=.7\columnwidth]{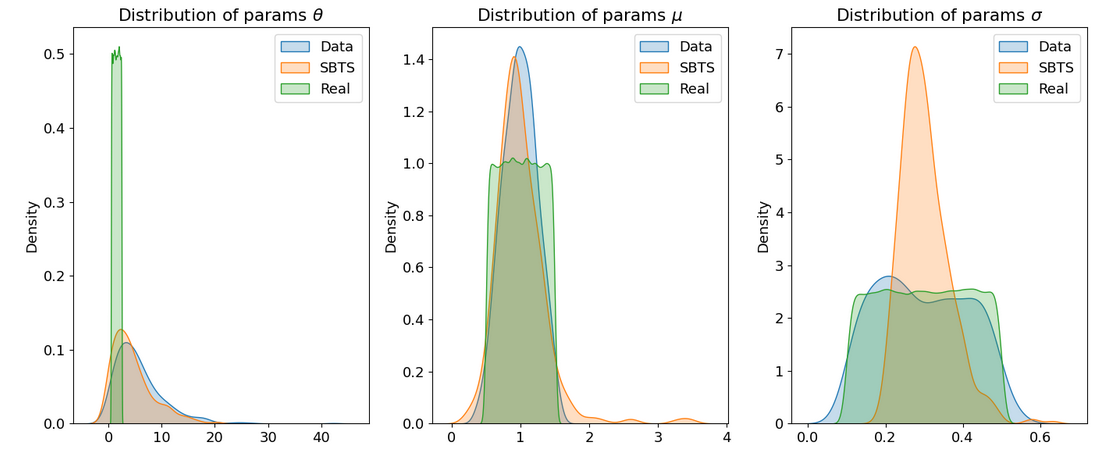}}
		\caption{Distribution of estimated Ornstein-Uhlenbeck parameters using MLE. We show in orange, blue and green the density respectively from the SBTS samples, data samples, and real range.}
		\label{fig:ou_params}
	\end{center}
\end{figure}

For the Ornstein-Uhlenbeck data, our results presented in \autoref{fig:ou_params} show that the estimated parameters are remarkably consistent between the real and synthetic data, with similar distributions observed for both.

\begin{figure}[h]
	\begin{center}
		\centerline{\includegraphics[width=.6\columnwidth]{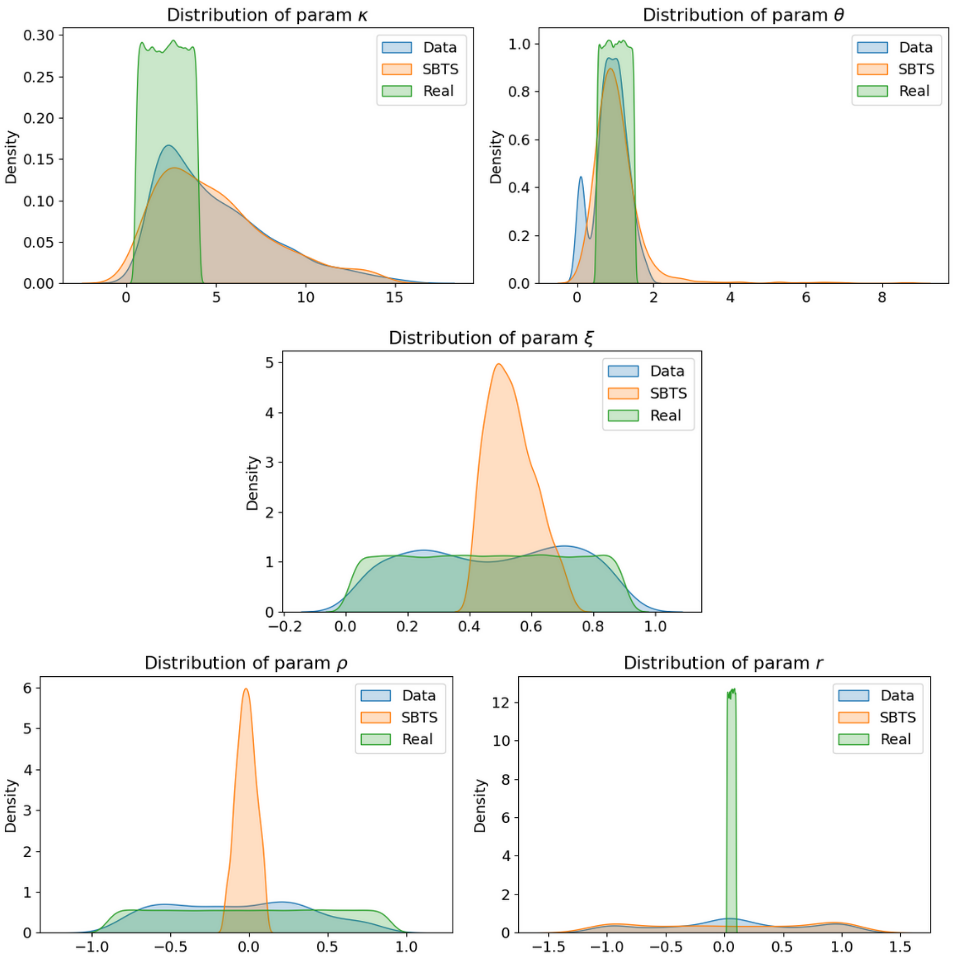}}
		\caption{Distribution of estimated Heston parameters using MLE. We show in orange, blue and green the density respectively from the SBTS samples, data samples, and real range.}
		\label{fig:heston_params}
	\end{center}
	\vskip -0.2in
\end{figure}

Regarding the Heston model, \autoref{fig:heston_params} shows that it yields similar results for $\kappa, \theta$ and $r$ that were consistent between real and synthetic data, while the estimated parameters $\xi$ and $\rho$ showed significant discrepancies. For the latter ones, SBTS exhibits a notable discrepancy, with the synthetic data yielding a Gaussian distribution with lower variance centered around the midpoint of the range used for random sampling. This suggests that the generative model tends to produce an averaged value for $\sigma$, rather than capturing the full range of variability present in the real data.

These findings indicate that SBTS may be less effective at capturing the nuances of volatility-related parameters, and highlights the need for further refinement to improve the robustness of the  model. This can be attributed to the inherent assumption of constant variance in the generative model, which is not a valid assumption for Heston processes that exhibit stochastic variance. However, it is worth noting that the parameter ranges were intentionally chosen to be extremely wide in order to assess how well SBTS can recover them, but in practice, such extreme values are rarely observed in real financial time series.

We repeated the experiment but using the same fixed parameters for each sample, rather than randomly sampling from a given range. Notably, this approach yielded consistent results between the real and synthetic data for all parameters, including the volatility-related ones (see \autoref{sec:appendix}). This reinforces the idea that SBTS is capable of accurately capturing the underlying dynamics when the parameters are fixed and consistent with the real data.

\section{Scaling Procedure in our Experiments}
\label{sec:data_scaling}

In this section, we discuss the scaling strategy employed in our experiments. Given the SDE governing the process:

\[
dX_t = \alpha_t^* dt +  dW_t^{\mathbb{P}},
\]

with $\alpha^*$ defined in \eqref{eq:a_star_equation}, the theoretical variance of this process is expected to be $\Delta t$. However, we observe that the model successfully generates data with volatility different from the expected one. This behavior can be attributed to the expression of $\alpha^*$ in equation \eqref{eq:a_star_equation}, which acts as a corrective term adjusting the path’s volatility if it deviates significantly from the expected level.

Two key terms contribute to this effect:  $\exp \left( - \frac{\lVert x_{i+1} - x \rVert^2}{2(t_{i+1} - t)} \right)$ and the indicator function in $\tilde{K}_i^m$. Furthermore, if the time series is not stationary, for instance when using raw prices instead of log-returns, the second term $\exp \left( \frac{\lVert x_{i+1} - x_i \rVert^2}{2(t_{i+1} - t_i)} \right)$ in the function $F_i$, would have a negligible weight, as $\lVert x_{i+1} - x_i \rVert \simeq 0$ for small $\Delta t$.

\begin{figure}[h]
	\begin{center}
		\centerline{\includegraphics[width=1\columnwidth]{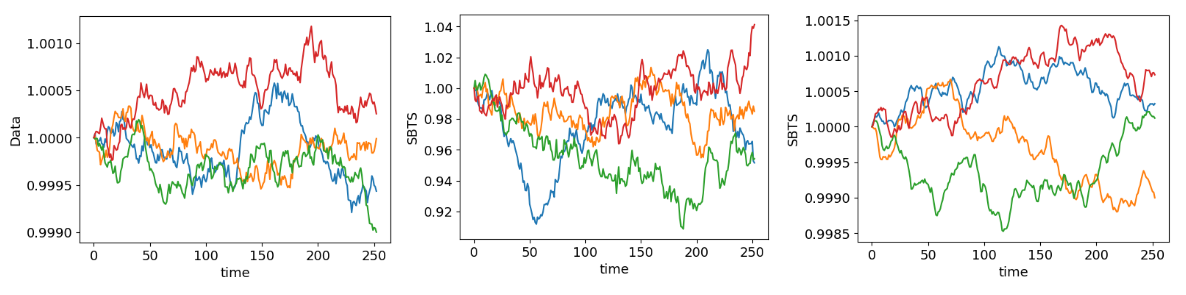}}
		\caption{Generation of an Ornstein-Uhlenbeck process with $\sigma=0.001, \theta \!=\! \mu \!=\! 1$. The unscaled SBTS model (\textit{middle}) fails to accurately capture the true variance of the process, whereas the scaled SBTS model (\textit{right}) better reproduces the expected variability. We used here $\Delta t = \frac{1}{252}$, $h=0.2$ and $k=1$.}
		\label{fig:scaled_data}
	\end{center}
\end{figure}

When the variance of the observed data significantly deviates from $\Delta t$, the drift term $\alpha_t^*$ is unable to properly correct the volatility of the generated paths, as illustrated in figure~\ref{fig:scaled_data}.

To address this issue, one possible approach would be to reduce $\Delta t$ significantly. However, this is not an optimal solution, as it introduces an additional hyperparameter and may distort the actual temporal frequency of the data. Instead, a more appropriate solution consists of rescaling the log-returns $R$ as follows:

\[
\tilde{R}_{t_1:t_N} = R_{t_1:t_N} \times \frac{\sqrt{\Delta t}}{\sigma(R_{t_1:t_N})},
\]

where $R_{t_1:t_N} = (R_{t_1}, \cdots, R_{t_N}) \in (\mathbb{R}^d)^N$ and $\sigma(R_{t_1:t_N}) \in \mathbb{R}^d$ denotes the empirical standard deviation of the data. To recover the original scale, one simply needs to multiply the generated log-returns by $\frac{\sigma(R_{t_1:t_N})}{ \sqrt{\Delta t}}$.

This transformation ensures that the variance of the scaled increments matches the theoretical variance of the process, thereby improving the stability and performance of the model in generating realistic paths.

\section{Conclusion and Future Work}
\label{sec:ccl}

In this study, we have demonstrated the effectiveness of the SBTS approach in generating high-quality  time series data. Our results show that SBTS consistently outperforms GAN approaches and achieves performance comparable to  state-of-the-art diffusion models. A key advantage of SBTS lies in its simplicity - it requires no pre-training, involves minimal parameter tuning, and enables fast data generation without requiring significant computational power, making it an appealing choice for time series synthesis across various domains.
 
Despite these strengths, we have also identified certain limitations of the SBTS framework. First, the kernel-based approach used to approximate the drift is highly sensitive to the choice of kernel bandwidth, which can hinder 
performance, especially when generating long time series. However, we have shown that introducing an assumption of finite Markovianity order effectively mitigates this issue without compromising the quality of the generated data.  Second, the current SBTS model assumes  constant variance, which may be insufficient for accurately modeling time series with stochastic volatility,  a common feature in financial data. 
Addressing this limitation is the focus of our ongoing work, where we are actively enhancing the SBTS framework by integrating stochastic variance, making it more powerful in capturing the complexity of real-world time series. Finally, it is well known that kernel-based methods struggle with high-dimensional data due to the curse of dimensionality. Estimating the drift (and volatility) using neural networks remains a challenging problem that is currently being addressed.


\bibliographystyle{ACM-Reference-Format}
\bibliography{sample-base}

\appendix

\section{Dataset}
\label{sec:appendix_dataset}

We provide a detailed description of the toy dataset used in this study:

\begin{itemize}
\item \textbf{GARCH: } In Figure~\ref{fig:h_selection} and~\ref{fig:demo}, we used the same GARCH model as defined in \cite{hamdouche2023generativemodelingtimeseries} :

$
\left\{
\begin{array}{ll}
	X_{t_{i+1}} = \sigma_{t_{i+1}}\epsilon_{t_{i+1}}  \\
	\sigma_{t_{i+1}}^2 = \alpha_0 + \alpha_1 X_{t_i}^2 + \alpha_2 X_{t_{i-1}}^2
\end{array}
\right.
$
with $\alpha_0=5, \alpha_1=0.4, \alpha_2=0.1$,  and $\epsilon_{t_i} \sim \mathcal{N}(0, 0.1), i = 1, \cdots, N$ are i.i.d.

\item \textbf{Sine: } As defined in \cite{NEURIPS2019_c9efe5f2}, we simulate multivariate sinusoidal sequences of different frequencies $\eta$ and phases $\theta$, providing continuous-valued, periodic, multivariate data where each feature is independent of others. For each dimension $i \in \{1, \cdots, 5\}, x_i(t) = \sin(2\pi\eta t + \theta)$, where $\eta \sim \mathcal{U}[0, 1]$ and $\theta \sim \mathcal{U}[-\pi, \pi]$.

\item \textbf{AR: } As defined in \cite{NEURIPS2019_c9efe5f2}, we simulate autoregressive multivariate Gaussian models such that $x_t = \phi x_{t-1} + Z, Z \sim \mathcal{N}(\mathbf{0},\sigma \mathbf{1} + (1-\sigma) \mathbf{I})$. We used $\phi = 0.5$, $\sigma=0.8$, $x_0 = 0$.
\end{itemize}

Moreover, it is noteworthy that for the datasets Stocks and Air, we first generated the log returns  before inverting them back to their base one scale, while we applied a standard normalization for Energy, as it proved to be more effective in our experiments. For all datasets, we employed a sample length of 24, except for the Multi-Stock and fBM datasets, where the sample length was set to 10 to maintain consistency with the values reported in the referenced papers.

\section{Implementation details of the evaluation metrics}
\label{sec:score}

All experiments were run on a system with an Intel Core Processor (Broadwell, no TSX, IBRS), 12 CPU cores, 1 thread per core, and a CPU clock speed of 2.6 GHz, except for the metrics related to the discriminative and predictive scores, which were computed using a single NVIDIA A100 SXM4 40 GB GPU.

\subsection{Table \ref{tab:real_data} and \ref{tab:toy_data} datasets}

In order to ensure comparable results, we employed the exact same code to compute both the discriminative and predictive scores, as detailed in \cite{NEURIPS2019_c9efe5f2}. For both metrics, we utilized a unidirectional GRU network. The discriminative score was computed on the inverted data in base one scale, whereas the predictive score was computed on these data after applying min-max scaling to ensure consistency with SOTA methods. 

The predictive score is computed by training a GRU network on the first $d-1$ features from $t_1$ to $t_{N-1}$ to predict the $d$-th feature at time steps $t_2$ to $t_N$, with the mean absolute error (MAE) evaluated over the entire predicted sequence. This approach leverages the assumption that if the generative model is well-trained, the first $d-1$ features contain meaningful information about the $d$-th feature due to inherent correlations, allowing for its accurate prediction. The model is trained exclusively on synthetic data and evaluated on real data.

For both scores, the GRU was trained on $3,000$ synthetic samples and an equal number of randomly selected real samples. The training was conducted over $2,000$ epochs with a batch size of $128$. The hidden dimension was set to $4$ for the discriminative score and to $\max(\frac{d}{2}, 1)$ for the predictive score, while the number of layers was set to $2$ and $1$, respectively.

To assess performance, we compared SBTS with GAN-based and diffusion-based models. For each dataset, we performed the test 10 times and report the mean score along with the standard deviation, calculated over these 10 test runs. Additionally, we incorporated SOTA results from existing papers, except for the Air and AR datasets with the ImagenTime model, as corresponding results were not found in the literature.

\subsection{Table \ref{tab:additional_results} datasets}

For the experiments on the Multi-Stock and fBM datasets, we utilized the same code as in \cite{tao2024highrankpathdevelopment}, performing 5 runs per test, in line with the setup used in the original paper. For the Multi-stock dataset, we have generated $1000$ time series with $h=0.2$ and $\Delta t_i = \frac{1}{252}$, and $10000$ time series for the fBM dataset with $h=0.2$ and $\Delta t_i = 1/10$.

\section{Robustness test}
\label{sec:appendix}

\begin{figure}[h]
	\begin{center}
		\centerline{\includegraphics[width=.8\columnwidth]{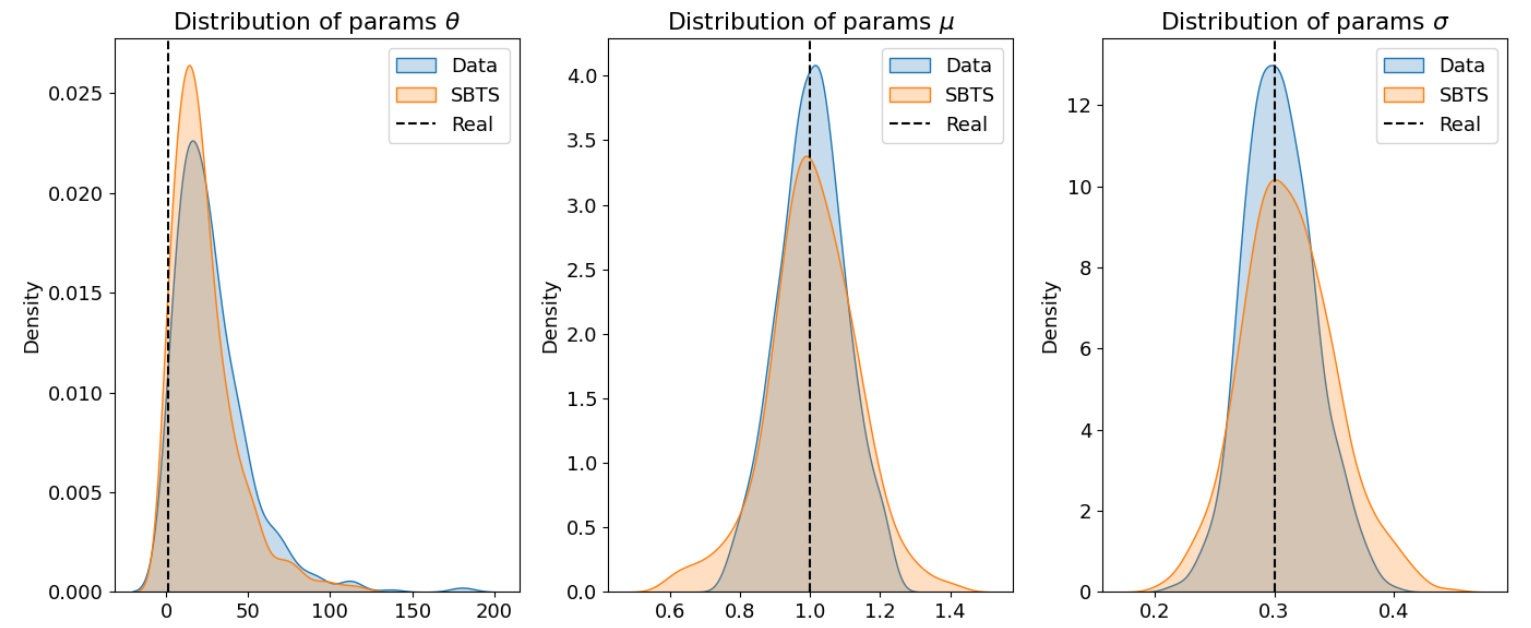}}
		\caption{Distribution of estimated Ornstein-Uhlenbeck parameters using MLE for fixed parameters. The orange and blue densities correspond to the SBTS samples and data samples, respectively, while the black line represents the true parameter.}
		\label{fig:ou_params2}
	\end{center}
	\vskip -0.2in
\end{figure}

\begin{figure}[h]
	\begin{center}
		\centerline{\includegraphics[width=.5\columnwidth]{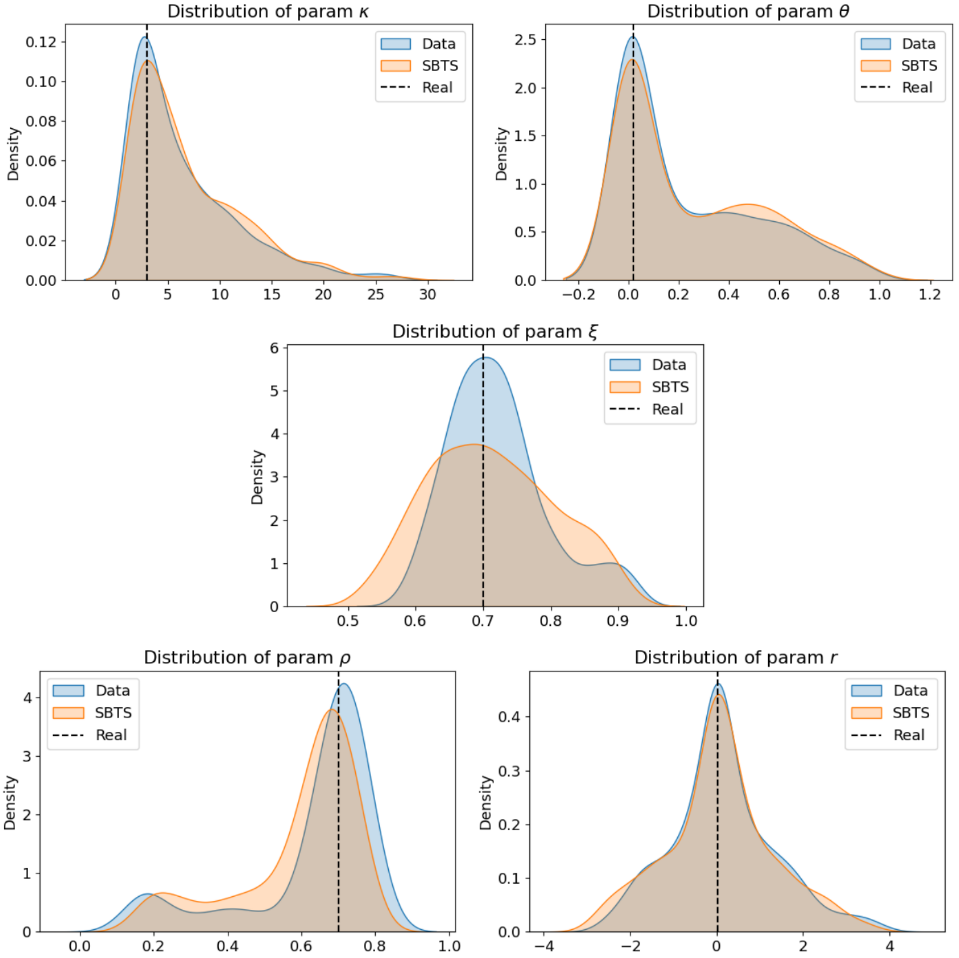}}
		\caption{Distribution of estimated Heston parameters using MLE for fixed parameters. The orange and blue densities correspond to the SBTS samples and data samples, respectively, while the black line represents the true parameters.}
		\label{fig:heston_params2}
	\end{center}
	\vskip -0.2in
\end{figure}

We present in figures~\ref{fig:ou_params2} and~\ref{fig:heston_params2} the distribution of the estimated parameters when they are fixed for all samples.

In addition, we present in Table \ref{tab:params_combined_resized} the parameter settings we employed in our experiments.

\begin{table}[h]
    \centering
    \caption{Parameter settings for the Ornstein-Uhlenbeck (left) and Heston (right) processes.}
    \label{tab:params_combined_resized}
    \resizebox{\columnwidth}{!}{%
    \begin{tabular}{@{}lccc|ccccc@{}}
        \toprule
        & $\theta$ & $\mu$ & $\sigma$ 
        & $\kappa$ & $\theta$ & $\xi$ & $\rho$ & $r$ \\
        \midrule
        Range 
        & [$0.5$, $2.5$] & [$0.5$, $1.5$] & [$0.1$, $0.5$]
        & [$0.5$, $4$] & [$0.5$, $1.5$] & [$0.01$, $0.9$] & [$-0.9$, $0.9$] & [$0.02$, $0.1$] \\
        Fixed 
        & $1.5$ & $1.0$ & $0.3$ 
        & $3.0$ & $0.5$ & $0.7$ & $0.7$ & $0.02$ \\
        \bottomrule
    \end{tabular}%
    }
\end{table}

Finally, we set the time step size to $h=0.6$ for the Ornstein-Uhlenbeck time series of length $252$ and $h=0.4$ for the Heston time series of length $100$, both with $N^{\pi}=200$, the Euler time steps between $t_i$ and $t_{i+1}$ and $k=1$. The time required to generate $1000$ samples was $659$ seconds and $548$ seconds, respectively, using Numba acceleration packages.

\end{document}